\documentclass{article}

\PassOptionsToPackage{numbers, compress}{natbib}
 \usepackage[preprint]{neurips_2025}

\usepackage{graphicx}
\usepackage{subcaption}
\usepackage{booktabs} % for professional tables
\usepackage{multirow}
\usepackage[utf8]{inputenc} % allow utf-8 input
\usepackage[T1]{fontenc}    % use 8-bit T1 fonts
\usepackage{hyperref}       % hyperlinks
\usepackage{url}            % simple URL typesetting
\usepackage{booktabs}       % professional-quality tables
\usepackage{amsfonts}       % blackboard math symbols
\usepackage{nicefrac}       % compact symbols for 1/2, etc.
\usepackage{microtype}      % microtypography
\usepackage{xcolor}         % colors
\usepackage{amsmath}
\usepackage{amssymb}
\usepackage{mathtools}
\usepackage{amsthm}
\usepackage[capitalize,noabbrev]{cleveref}
\usepackage{xspace}
\usepackage{placeins}
\usepackage{float}
\usepackage{wrapfig}

\newcommand{\netq}[1]{\mathcal{E}^Q_\theta(x_{#1}, #1)}
\newcommand{\epsq}{\mathcal{E}^Q_\theta}
\newcommand{\net}[1]{\mathcal{E}_\theta(x_{#1}, #1)}
\newcommand{\eps}{\mathcal{E}_\theta}
\newcommand{\qsched}{\texttt{Q-Sched}\xspace}

\title{$\texttt{Q-Sched}$: Pushing the Boundaries of Few-Step
Diffusion Models with Quantization-Aware Scheduling}

\author{%
  Natalia Frumkin \\
  The University of Texas at Austin\\
  \texttt{nfrumkin@utexas.edu} \\
  \And
  Diana Marculescu \\
  The University of Texas at Austin\\
  \texttt{dianam@utexas.edu} \\
}

\begin{document}

\maketitle
\begin{figure}[H]
    \vspace{-20pt}
    \centering    \includegraphics[width=\linewidth]{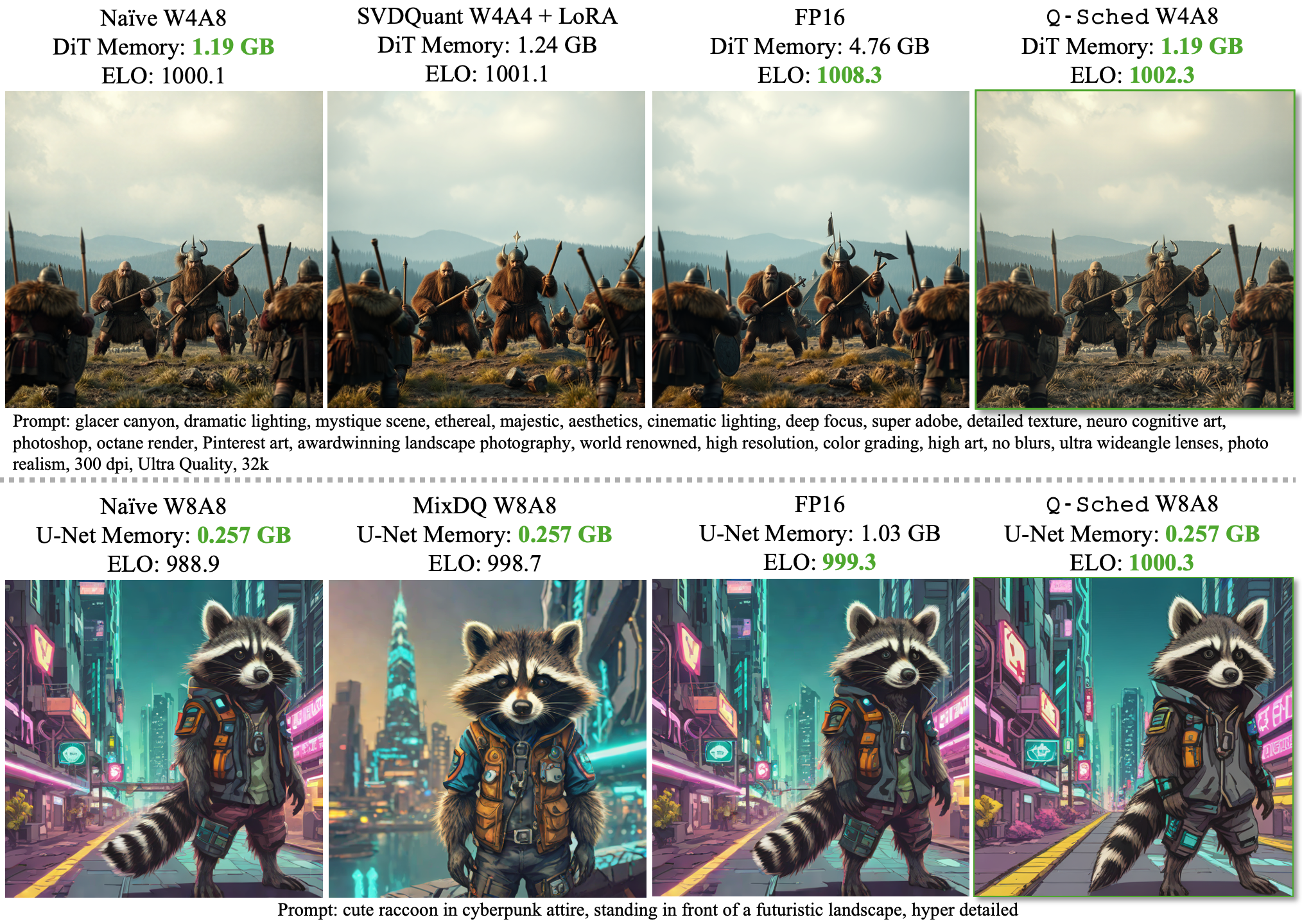}
    \vspace{-15pt}
    \caption{
    	\looseness=-1
        \qsched introduces a quantization-aware noise scheduler to few-step diffusion backbones and achieves excellent image fidelity. We find quantization and few-step diffusions to be complementary model compression strategies.
    }
    \vspace{-10pt}
    \label{teaser}
\end{figure}

\begin{abstract}
Text-to-image diffusion models are computationally intensive, often requiring dozens of forward passes through large transformer backbones. For instance, Stable Diffusion XL generates high-quality images with 50 evaluations of a 2.6B-parameter model, an expensive process even for a single batch. Few-step diffusion models reduce this cost to 2-8 denoising steps but still depend on large, uncompressed U-Net or diffusion transformer backbones, which are often too costly for full-precision inference without datacenter GPUs. These requirements also limit existing post-training quantization methods that rely on full-precision calibration.

We introduce Q-Sched, a new paradigm for post-training quantization that modifies the diffusion model scheduler rather than model weights. By adjusting the few-step sampling trajectory, Q-Sched achieves full-precision accuracy with a $4\times$ reduction in model size. To learn quantization-aware pre-conditioning coefficients, we propose the JAQ loss, which combines text-image compatibility with an image quality metric for fine-grained optimization. JAQ is reference-free and requires only a handful of calibration prompts, avoiding full-precision inference during calibration.

Q-Sched delivers substantial gains: a 15.5\% FID improvement over the FP16 4-step Latent Consistency Model and a 16.6\% improvement over the FP16 8-step Phased Consistency Model, showing that quantization and few-step distillation are complementary for high-fidelity generation. A large-scale user study with more than 80,000 annotations further confirms Q-Sched’s effectiveness on both FLUX.1[schnell] and SDXL-Turbo. Code will be released upon publication.
Our code is available at \href{https://github.com/enyac-group/q-sched}{https://github.com/enyac-group/q-sched}.
\end{abstract}

\section{Introduction}
\label{intro}
Diffusion models are a powerful class of deep generative models with impressive results across computer vision~\cite{amit2021segdiff, baranchuk2021label, brempong2022denoising, ho2022cascaded, meng2021sdedit, yang2022diffusion}, natural language processing~\cite{austin2021structured, li2022diffusion}, multi-modal modeling~\cite{avrahami2022blended, ramesh2022hierarchical}, and interdisciplinary fields~\cite{anand2022protein, cao2022high}. State-of-the-art systems like Stable Diffusion XL~\cite{podell2023sdxl, meng2021sdedit} and CogVideoX~\cite{yang2024cogvideox} demand server-grade GPUs and significant computational resources at inference time. This is due to the expensive denoising process, which often requires 40–1,000 steps per image or video, with each step invoking a deep, transformer-based noise estimation network.

To accelerate inference, two main strategies are typically considered: (1) reducing the number of function evaluations (\textit{i.e.}, denoising steps), and (2) lowering the cost per evaluation. The latter can be addressed using standard model compression techniques such as quantization~\cite{he2024ptqd, guo2022squant}, pruning~\cite{fang2024structural}, or knowledge distillation~\cite{huang2024knowledge}. However, decreasing the number of denoising steps requires careful redesign of the sampling procedure, as naive reductions can severely degrade output quality.

\begin{figure}
  \centering
  \begin{subfigure}[t]{0.6\textwidth}
    \centering
    \includegraphics[height=0.64\linewidth]{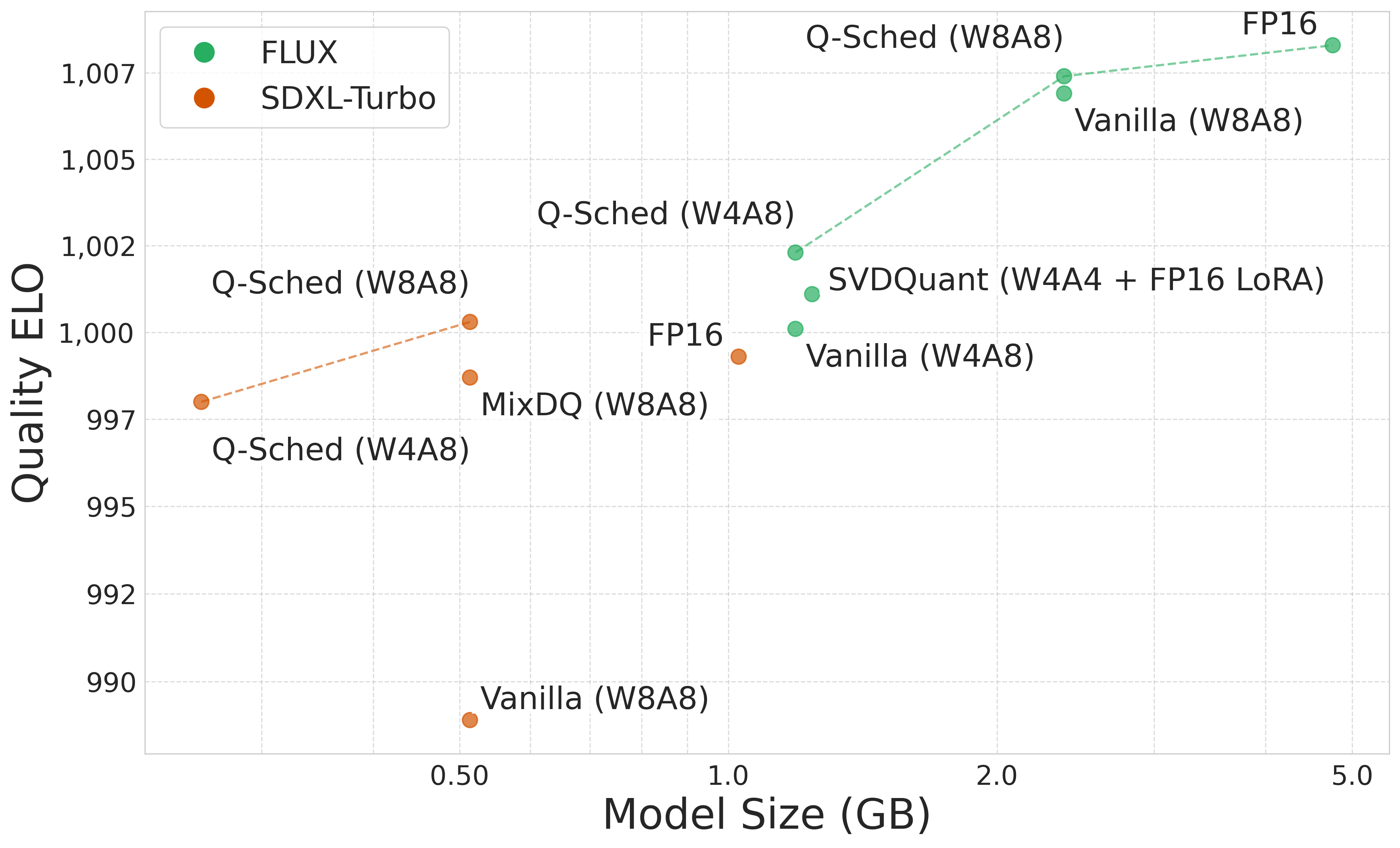}
    \caption{ELO Score \textit{vs.} Model Size for various quantization methods on FLUX.1[schnell]~\cite{flux1schnell} and SDXL-Turbo~\cite{sauer2024adversarial}.}
    \label{fig:pareto}
  \end{subfigure}
  \hfill
  \begin{subfigure}[t]{0.38\textwidth}
    \centering
    \includegraphics[height=\linewidth]{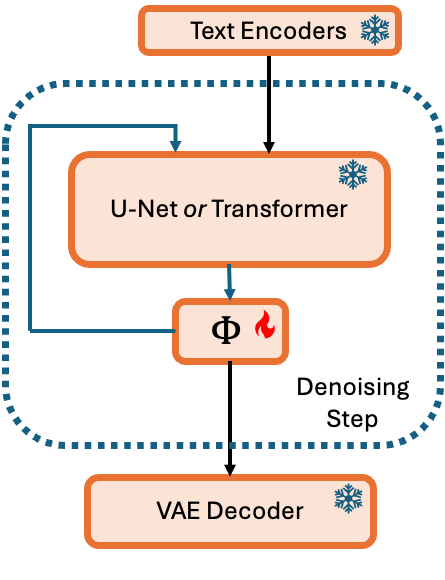}
    \caption{\qsched optimizes the few-step diffusion scheduler, making it highly modular quantization method.}
    \label{fig:pipeline}
  \end{subfigure}
  \caption{\qsched's quantization-aware scheduling enables state-of-the-art image fidelity across multiple compressed few-step models.   \qsched directly optimizes the few-step diffusion's scheduler (see \cref{fig:pipeline}) whereas prior work directly optimizes the transformer or U-Net backbone.}
  \label{fig:sidebyside}
\end{figure}

\begin{figure*}[t]
    \centering
    \includegraphics[width=0.75\linewidth]{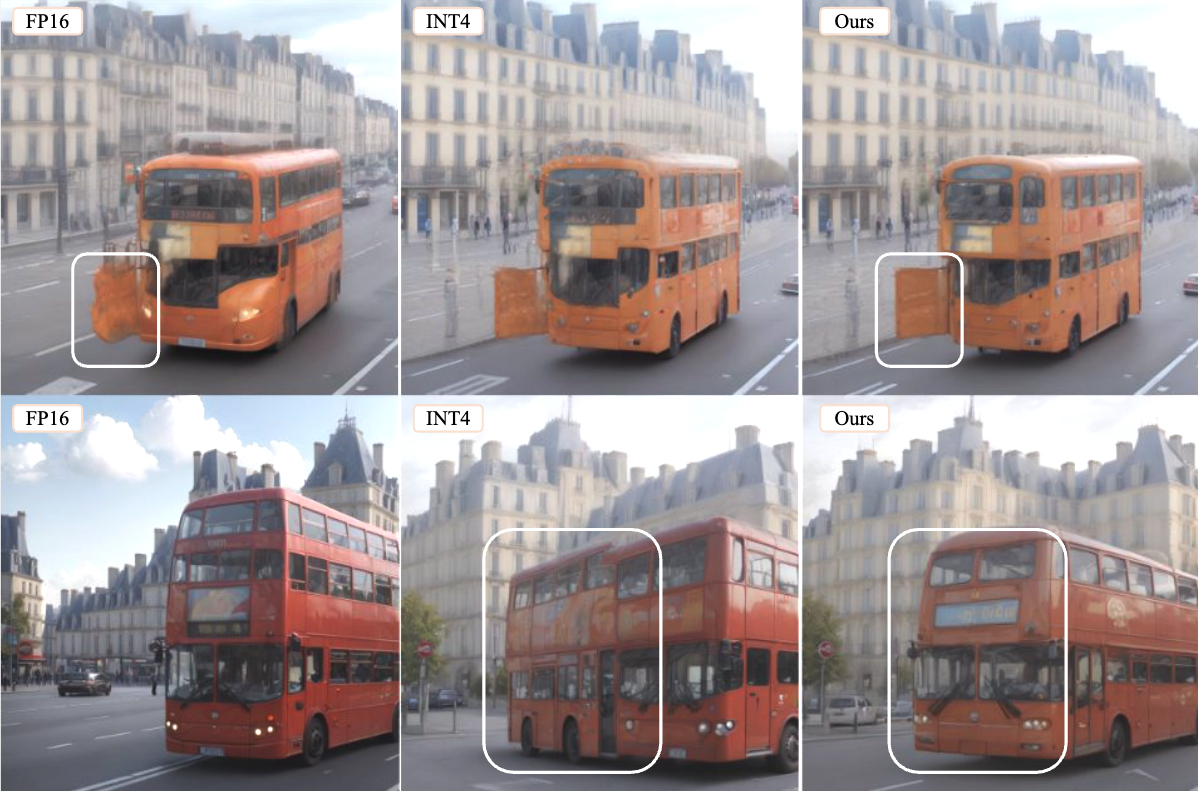}
    \caption{4-Step (top row) and 8-Step (bottom row) LCMs. Prompt: "a car and a bus on a french highway". \qsched is capable of avoiding artifacts present in the FP16 or INT4 generative images. \qsched is close to the original schedule since it generates similar images yet our optimized schedule allows for \qsched to avoid some artifacts generated from the original schedule. }
    \label{fig:fig1_lcm}
\end{figure*}

Few-step diffusion models~\cite{song2023consistency, yin2023one, luo2023latent} reduce the number of steps taken along the sampling trajectory of a diffusion model's ordinary or stochastic differential equation (ODE/SDE). These methods, along with a variety of other diffusion models, rely on an accurate probability flow ODE or variance-preserving SDE to express the relationship between the noise estimation network and the final generated image~\cite{song2021scorebased}. However, if the noise estimation network becomes corrupted, such as by compression, it can significantly change the ODE/SDE trajectory. 

This work presents a novel quantization-aware noise schedule, dubbed \qsched, which modifies the few-step diffusion's sampling trajectory to improve performance of few-step model.

Our contributions are summarized as follows:

\begin{enumerate}
    \item We present a novel quantization-aware scheduler that can be applied to existing few-step diffusion models. As shown in \cref{fig:attn_grabber}, \qsched is capable of outperforming the original few-step diffusion, achieving an impressive \textbf{15.5\% FID improvement} over an existing 4-step LCM \textit{while} reducing model size.
    \item \qsched's novel preconditioning coefficients enable the quantized model to diverge from the potentially overfit few-step diffusion model and bypass artifacts created from the distillation process and quantization.

    \item We propose the Joint Alignment-Quality (JAQ) loss, a reference-free metric that balances perceptual fidelity and text-image alignment. JAQ enables fine-grained control over visual attributes (\textit{e.g.}, texture, detail, saturation) without requiring access to the full-precision model. 
    
    \item We perform more than 80,000 human preference annotations and find that \qsched outperforms MixDQ~\cite{zhao2024mixdq} on SDXL-Turbo and SVDQuant~\cite{lisvdquant} on FLUX.1 in terms of perceived image quality.
\end{enumerate}

In \cref{teaser}, we illustrate that \qsched can achieve the best ELO rating when stacked up against other images in a pair-wise image quality comparison. Furthermore, we show in\cref{fig:pareto} that \qsched is Pareto optimal with respect to both ELO and model size. These results highlight \qsched's ability to balance perceptual quality and efficiency better than competing methods.

\section{Background and Related Work}

Diffusion models estimate an unknown data distribution $p_{data}(\mathbf{x})$ through a perturbation scheme known as denoising~\cite{ho2020denoising}. In this setup, we consider a set of sequential Gaussian perturbations which transform an image $x_0$ to random noise $x_T$. The forward process involves perturbing an image to noise, and the reverse process (denoising) involves removing noise from $x_T$ until we obtain $\hat{x}_0$. As shown by \cite{song2021scorebased}, the \textit{continuous-time} denoising process can be modeled as the solution to the following stochastic differential equation:
\begin{equation}
    d\mathbf{x_t} =  \mathbf{f}(\mathbf{x_t}, t) d\textit{t}+g(t) d \mathbf{w} 
\end{equation}
where $\mathbf{f}(\cdot, \cdot): \mathbb{R}^d\times \mathbb{R} \rightarrow\mathbb{R}^d$ and $g(\cdot): \mathbb{R}\rightarrow\mathbb{R}$ correspond to the drift and diffusion coefficients. $\mathbf{w}$ is the Weiner process and $t \in [0,T]$ where T is a positive constant representing the number of diffusion steps. In order to generate $\hat{x_0}$ from the denoising process, we sample from the probability flow ODE using standard numerical methods (Euler~\cite{song2021scorebased}, Heun~\cite{karras2022elucidating}, \textit{etc.}). The probability flow ODE is defined as:
\begin{equation}
   d\mathbf{x_t} = \Bigg[ \mathbf{f}(x_t, t) - \frac{1}{2}g(t)^2 \nabla_x\log p_t(\mathbf{x_t},\sigma_t)\Bigg]d\textit{t}
   \label{eq:prob_flow}
\end{equation}

where $\nabla_x \log p_t(x_t, \sigma_t)$ is the Stein's score. In the case of EDM~\cite{karras2022elucidating} and Consistency Models~\cite{song2023consistency}, the drift and diffusion coefficients are set to $\mathbf{f}(x_t,t) = 0$ and $g(t) = \sigma_t = \sqrt{2t}$. It follows from these design choices that $p_t(x_t, \sqrt{2t}) = p_{data}(x) \bigotimes \mathcal{N}(0, t^2I)$ where $\bigotimes$ is the convolution operator.

The Stein's score is learned through denoised score matching using a denoiser $D_\theta$:

\begin{equation}
    \nabla_x \log p_t(x_t, \sigma_t) = \frac{D_\theta(x_t, \sigma_t) - x_t}{\sigma_t^2}
    \label{eq:denoiser}
\end{equation}

which is parametrized by a neural network $\eps$. A common preconditioning scheme developed by EDM~\cite{karras2022elucidating} is:
\begin{equation}
    D_\theta(\mathbf{x_t}, \sigma_t) = c_{skip}(\sigma_t)\mathbf{x_t} + c_{out}(\sigma_t)\net{t} \bigg(c_{in}(\sigma_t)\mathbf{x_t}, c_{noise}(\sigma_t)\bigg)
    \label{eq:precond}
\end{equation}

where $c_{skip}(\cdot)$, $c_{out}(\cdot)$, $c_{in}(\cdot)$, and $c_{noise}(\cdot)$ are hyperparameters chosen prior to training. The DDIM~\cite{song2020denoising} and Stable Diffusion~\cite{rombach2022high} preconditioning is slightly different and will be described in more detail when discussing Consistency Models. In the following section, we will present a common method for compressing diffusion models.

\paragraph{Few-Step Diffusion Models}
\label{sec:background}
% \paragraph{Compression for Text-to-Image Generative Models}
Distillation is a leading approach for compressing diffusion models by reducing denoising steps. These \textit{few-step} models offer substantial inference speedups with acceptable image quality, making them ideal for large-scale deployment. For instance, Instaflow~\cite{liu2023instaflow} distills StableDiffusion v1-4 from $T=50$ to $T=1$, cutting inference time from 2.9s to 0.09s—enabling servers to handle orders of magnitude more users.

Consistency Models~\cite{song2023consistency} use a consistency function to guide distillation—covered in the next section. Another popular approach, Rectified Flow, enables 1-step distillation by straightening the ODE trajectory, with Instaflow achieving this via coupling between noise and image spaces. Adversarial Diffusion Distillation (ADD)~\cite{sauer2024adversarial} further improves sample quality by introducing a discriminator that encourages the student model to generate outputs indistinguishable from the teacher's.

\paragraph{Consistency Models}
\label{subsec:cm}
In consistency models, the consistency function maps any point along the sampling trajectory to the final image $x_0$:

\begin{equation}
    x_0 = \mathcal{F}(x_t, t) \hspace{3mm} \forall t \in [0,T]
\end{equation}

which is equivalent to the self-consistency property:

\begin{equation}
    \mathcal{F}(x_t, t) = \mathcal{F}(x_s, s) \hspace{3mm}\forall t,s \in [0,T] .
\end{equation}

Consistency Model variants arise from how one parametrizes the consistency function $\mathcal{F}(x_t, t)$. Borrowing from EDM,  Latent Consistency Models (LCMs)~\cite{lcm} parameterize the consistency function using the same preconditioning as described in \cref{eq:precond}. LCMs have grown to be a mainstream generative model due to their impressive performance with modest computational costs. 

Following LCMs success, Trajectory Consistency Distillation (TCD)~\cite{tcd} developed a new trajectory consistency function to further generalize the self-consistency function. They parametrize the consistency function without using $c_{skip}, c_{out}$ and instead opt for the DDIM-style parametrization:

\begin{equation}
    \mathcal{F}_\theta(x_t, t) = \frac{\alpha_0}{\alpha_t}\mathbf{x_t} - \alpha_0 (\frac{\sigma_t}{\alpha_t} - \frac{\sigma_0}{\alpha_0}) \net{t} \hspace{1mm} .
\end{equation}

The TCD schedule is used in Phased Consistency Models (PCMs)~\cite{pcm}, which improve upon LCMs by adding an improved classifier-free guidance (CFG) solver, higher CFG controllability, along with better distillation stability. Our proposed \qsched scheduler can be applied to the preconditioning of both LCMs and PCMs to achieve highly distilled, quantized generative models.

\paragraph{Quantization for Diffusion Models}

There are a variety of methods applying model compression on standard text-to-image diffusion models. To compress the noise estimation network, $\eps$, 
PTQ4DM~\cite{shang2022post}, ADP-DM~\cite{wang2023towards} and Q-Diffusion~\cite{li2023q} use timestep aware calibration methods to generate a calibration set for post-training quantization (PTQ). These methods consider how to learn the optimal quantization scheme for a given network across the many function evaluations required during the forward process. 
Prior work~\cite{wang2024quest, wang2023towards, chen2024qdit} also addresses activation quantization since model activations change with each timestep.  Quantization schemes can be applied dynamically, such as using the TDQ module~\cite{so2024temporal}, to learn a set of quantization parameters for each timestep. These "timestep aware"  methods do not consider how quantization incurs a distribution shift at each timestep. 

To address distribution shifts, Q-DM~\cite{li2024q} considers how the noise function accumulates errors across time and proposes a noise-estimation scheme to reduce the quantization error. Similarly, PTQD~\cite{he2024ptqd} addresses the error accumulation across timesteps by applying bias correction directly to the sampled image $x_t$. PTQD models the distribution shift from full precision to quantized model using a linear approximation:

\begin{equation}
    \netq{t} = (1+\gamma) \cdot \eps + \delta 
\end{equation}

and aims to learn $\gamma$ by standard deviation correction and considers $\delta$ as uncorrelated quantization noise to be modelled as Gaussian. PTQD provides the most practical scheduler modification for quantized diffusions which can be adapted for few-step diffusions. We explain how to adapt PTQD to the TCD scheduler in \cref{sec:ptqd}. Next, we demonstrate how to apply \qsched to TCD and show how the same approach generalizes to other few-step samplers.

\section{Quantization-Aware Scheduling}

To prepare the TCD scheduler for optimization with \qsched, let us consider sampling with a quantized network. TCD's Strategic Stochastic Sampling (SSS)~\cite{tcd} using a quantized network $\netq{t}$ is given by:

\begin{equation}
\begin{aligned}
    \mathbf{x_s} = &\frac{\alpha_s}{\alpha_{s'}} \Big(\alpha_{s'}\frac{\mathbf{x_t} - \sigma_t \netq{t}}{\alpha_t} + \sigma_{s'}\netq{t} \Big) + \eta \mathbf{z}
\end{aligned}
\label{eq:sss}
\end{equation}

where a denoised image, $x_s$, is generated from a previous denoised image, $x_t$, at timestep $t$. The noise schedule is given by $\sigma, \alpha$ and the sampler injects stochastic noise sampled from a  distribution $\mathbf{z} \sim N(0,I)$. The stochastic control parameter, $\eta$, is defined as:

\begin{equation}
    \eta = \sqrt{1 - \frac{\alpha_s^2}{\alpha_{s'}^2}}
\end{equation}

and can be manually overridden during sampling to generate images with different levels of stochasticity. The TCD sampler in \cref{eq:sss} is a state-of-the-art few-step diffusion sampler which is used in Phased Consistency Models. Note that the schedule relies on two inputs from the previous time-step: $x_t$ and $\netq{t}$. These will be particularly relevant when applying \qsched.

\paragraph{\qsched: A Learnable Schedule Pre-Conditioner}

We introduce \qsched, a lightweight, post-training method that modifies the noise schedule of few-step diffusion models using two learnable scalar preconditioning coefficients: $c_x$ and $c_\epsilon$, applied to $x_t$ and the quantized noise prediction $\epsq$, respectively. As illustrated in \cref{fig:pipeline}, \qsched operates independently of the model backbone (U-Net or transformer), making it broadly compatible with any few-step diffusion model whose scheduler resembles the TCD scheduler.

Quantized diffusion models often suffer from artifacts such as distortions and hallucinations (\cref{fig:artifacts}). Traditional methods mitigate this by modeling distributional shifts from full-precision outputs using calibration data. In contrast, \qsched does not require access to the original model or any distribution statistics. Instead, it directly learns noise schedule corrections by optimizing a reference-free image quality metric, the JAQ loss.

The modified sampling procedure is:

\begin{equation}
    \mathbf{x_s} = \frac{\alpha_s}{\alpha_{s'}} \Big(\alpha_{s'}\frac{c_x\mathbf{x_t} - \sigma_t c_{\epsilon}\netq{t}}{\alpha_t} + \sigma_{s'}c_{\epsilon}\netq{t} \Big) 
+ \sqrt{1 - \frac{\alpha_s^2}{\alpha_{s'}^2}} \mathbf{z} 
\end{equation}

During distillation, few-step diffusion models compress the standard diffusion process into a couple of steps, removing the Gaussian denoising assumption and producing models that are sensitive to further model compression. We find that two coefficients are sufficient to learn a corrected sampling schedule which rivals its full precision counterparts for some few-step models. Our quantization-aware schedule is a new paradigm because it does not rely on any distribution or bias correction, but rather optimizes the preconditioning coefficients with respect to the JAQ loss, our proposed image quality metric. In using an image quality metric rather than a mean or standard deviation adjustment, our noise schedule has the ability to learn the best sampling trajectory rather than adhering to the full precision models trajectory, which for distilled few-step diffusions, is likely to be overfit.

To learn our preconditioning coefficients, $c_x, c_\epsilon$, we apply grid search and evaluate each coefficient combination with respect to the JAQ loss. Next, we will discuss our new reference-free loss function, JAQ, and its benefits over existing image assessment tools.

\paragraph{JAQ: A Joint Alignment Quality Loss Function}

Reference-free metrics such as CLIPScore~\cite{hessel2021clipscore} have become essential for quick evaluation of text-to-image generation models. Unlike FID~\cite{fid}, SSIM~\cite{ssim}, and other comparative metrics, reference-free metrics do not rely on a ground truth reference image and therefore are very useful in generative tasks when a ground truth is not available. When quantizing these generative models, an image generated from a quantized model, $x_q$, has an altered sampling trajectory from the original full precision model. This is evident in \cref{fig:fig1_lcm}, where $x_q$ produces a different, sometimes cleaner image than full precision. In short, the quantized model takes an altered sampling trajectory which coarsely follows the full precision model yet generates sufficient differences that reference-based metrics do not capture the image's detail.

Our Joint Alignment Quality loss combines a text-to-image compatibility score with a pure image quality score to achieve better results than simply optimizing with respect to metrics such as CLIPScore or CLIP-IQA~\cite{clipiqa} independently. We design the JAQ loss so that it can better differentiate between images that are highly similar to one another, whereas standard image quality metrics are designed to rank images that come from a much larger distribution. Given a text-to-image compability metric, $\texttt{TC}(x)$, and a pure image quality metric, $\texttt{IQ}(x)$, JAQ combines them as follows:

\begin{equation}
    \texttt{JAQ}(x) = \texttt{TC}(x) + k \cdot \texttt{IQ}(x)
    \label{eq:kscore}
\end{equation}

Only optimizing with respect to a text-to-image compatibility metric such as CLIPScore causes loss of image details as we only optimize for how closely the image follows the prompt (see \cref{fig:img_quality}). Furthermore, we find that CLIPScore lacks adequate sensitivity to image artifacts due to quantization (examples of artifact types are shown in \cref{fig:artifacts}).

Conversely, if we only use an image quality metric, our \qsched scheduler will optimize for generating details that may not otherwise be in the prompt. JAQ allows us to balance these two conflicting objectives using a linear combination with the parameter $k$ controlling the tradeoff between text fidelity and image detail.

\section{Experiments}

\paragraph{Experimental Setup}
We apply \qsched across diverse few-step diffusion models, including both U-Net~\cite{ronneberger2015u} and DiT~\cite{peebles2023scalable} backbones, as well as consistency (LCM~\cite{lcm}, PCM~\cite{pcm}) and flow-matching strategies. We further compare against recent state-of-the-art models such as SDXL-Turbo~\cite{sauer2024adversarial} and FLUX.1~\cite{flux1schnell}. We evaluate two regimes: 4W8A and 8W8A. Most models remain robust, with 4W8A outputs often matching full-precision quality. Due to metric limitations on high-fidelity images~\cite{jayasumana2024rethinking}, we primarily report FID for 4W8A. Only the U-Net or DiT backbone is quantized, as it dominates model size (\cref{tab:model_size}).

LCM and PCM are evaluated at 2, 4, and 8 denoising steps on COCO-30k~\cite{lin2014microsoft}, using FID (\textit{vs.} real images), \texttt{CLIPScore} (for prompt-image alignment), and FID-SD (\textit{vs}. Stable Diffusion outputs). FLUX.1 and SDXL-Turbo are assessed on the SVDQuant~\cite{lisvdquant, li2024playground} subset of MJHQ-30k, with FID and a human preference study for evaluation. MJHQ-30k is a collection of 5,000 high-quality Midjourney prompts from 10 common categories. We report FID for SDXL-Turbo and conduct a human preference study to fully evaluate these models.

We use two variants of the Joint Alignment Quality (JAQ) loss: one based on CLIP-derived metrics, the other on human preference scores. The CLIP-based version uses $\texttt{TC(x)} = \texttt{CLIPScore(x)}$ and $\texttt{IQ(x)} = \texttt{CLIP-IQA(x)}$. For SDXL-Turbo and FLUX.1, we apply a preference-based variant using $\texttt{TC(x)} = \texttt{AQ-MAP(x)}$ and $\texttt{IQ(x)} = \texttt{HPSV2(x)}$, where AQ-MAP~\cite{li2024qref} produces a spatial alignment score and HPSV2~\cite{wu2023human} is fine-tuned on real human preferences. We set $k=2$ in both cases.

\begin{table*}
\centering
\caption{Comparison of different schedulers on Phased Consistency Models and Latent Consistency Models using a Stable Diffusion v1-5 backbone. The original schedule is TCD~\cite{tcd} for Phased Consistency Models and the Multi-step Consistency Sampling~\cite{lcm} for Latent Consistency Models. The FID and CLIPScore are calculated with respect to the COCO-30k dataset. NFEs stands for number of function evaluations referring to the number of passes through the network $\netq{t}$.}
\begin{tabular}{c c c c | c  c | c  c }
\toprule
\multirow{2}{*}{NFEs}& \multirow{2}{*}{Precision} & \multirow{2}{*}{Schedule} & Calibration & \multicolumn{2}{c|}{PCMs}  & \multicolumn{2}{c}{LCMs} \\
 &   &   & Size & FID & CLIPScore & FID  & CLIPScore  \\
\midrule
\multirow{4}{*}{2}  & FP16  & Original & - & 24.17 &  \underline{25.489}  & 38.74 &    \underline{25.155} \\
& W4A8 & Original & - & 28.70 & 25.343 & 40.93 & 24.886 \\
& W4A8 & PTQD & 1024 & \underline{23.33} & 25.265 & \underline{37.59} & 24.919 \\
& W4A8 &  \qsched & 5 & \textbf{22.24}  & \textbf{25.543}  & \textbf{32.50} & \textbf{25.152} \\
\midrule
\multirow{4}{*}{4}  & FP16  & Original & - & 23.29  & 25.482 & \underline{31.94} & \textbf{25.969} \\
& W4A8 & Original & - & 23.08 & 25.557 & 38.41 & \underline{25.456} \\
& W4A8 & PTQD & 1024 & \underline{19.42} & \underline{25.639}  & 39.72 &  24.678 \\
& W4A8 &  \qsched & 5 & \textbf{17.39} & \textbf{25.715 }& \textbf{26.98} & 25.336 \\
\midrule
\multirow{4}{*}{8} & FP16 & Original & - & 20.15 & \textbf{25.714} & \underline{27.34}  & \textbf{26.052} \\
& W4A8 & Original & - & 18.48 & \underline{25.664} & 27.55 & \underline{25.397} \\
& W4A8 & PTQD & 1024 & \textbf{15.85} & 25.770 & 28.06 & 25.241    \\
& W4A8 & \qsched & 5 & \underline{16.83} & 25.698 & \textbf{25.82} &  25.214 \\
\bottomrule
\end{tabular}
\label{tab:main}
\end{table*}

\paragraph{Calibration Set} 
We compare our results to PTQD~\cite{he2024ptqd}, the only other quantization-aware noise scheduler designed for few-step diffusion. While PTQD relies on a 1,024-image calibration set generated from a full-precision model, our approach requires only a small set of representative prompts. Specifically, we hand-curate calibration prompts for the latent and phased consistency models, and sample from the sDCI prompt set~\cite{lisvdquant} for SDXL-Turbo and FLUX.1 [schnell]. Our compact 20-image calibration set is reused across multiple evaluations, as larger sets would be prohibitively costly in post-training quantization.

\paragraph{Results: Latent and Phased Consistency Models}

In \cref{fig:fig1_lcm}, we consider three different quantized noise schedules and their performance on two consistency model families. We show that \qsched is able to overcome some artifacts present in the full precision and quantized models, illustrating that it learns a new few-step sampling trajectory which rivals the original model. Our method can produce greater detail than both FP16 and 4W8A ( see \cref{fig:good_captions} ) for a variety of images. Furthermore, \qsched, provides excellent FID results in comparison to full precision. This illustrates that modifying the few-step diffusion's noise schedule can allow the diffusion model to generate \textit{better} images than the full precision model. \qsched outperforms PTQD noise schedules in 4/6 consistency model variants on StableDiffusion v1-5 with a fraction of the calibration set (see \cref{tab:main}).

PTQD performs correction to the quantized model's distribution \textit{with respect to its full precision counterpart}. In contrast, \qsched does not rely on images from a full precision model and can be adjusted to outperform a flagship few-step diffusion model with only five prompts.  As consistency models are distilled from a large diffusion model, we'd expect them to be very sensitive to quantization as often distilled models are much more sensitive to compression. Surprisingly, the opposite is true. \qsched can outperform a full precision few-step diffusion model by \textbf{16.1\%, 15.5\%, 5.6\%} for a \textbf{2-step, 4-step, and 8-step model} respectively, illustrating that quantization and few-step distillation are complementary model compression techniques.

\begin{table}[htbp]
  \centering
  \begin{subtable}[t]{0.5\textwidth}
    \centering
    \small
    \begin{tabular}{c c | c c c}
      \toprule
      Scheduler & Precision & FID & FID-SD & CLIPScore \\
      \midrule
      TCD & FP16 & \textbf{18.65} & \textbf{10.45} & \textbf{26.531} \\
      \midrule
      TCD & W4A8 & 22.70 & 12.51 & 26.241 \\
      PTQD & W4A8 & 161.96 & 176.29 & 25.910 \\
      \qsched & W4A8 & \underline{18.89} & \underline{12.17} & \underline{26.513} \\
      \bottomrule
    \end{tabular}
    \caption{Comparison on a 2-step Phased Consistency model using the Stable Diffusion XL backbone. FID-SD is computed relative to images generated by Stable Diffusion XL using corresponding COCO-30k prompts.}
    \label{tab:sdxl}
  \end{subtable}
  \hfill
  \begin{subtable}[t]{0.4\textwidth}
    \centering
    \begin{tabular}{l c | c}
      \toprule
      Method & Precision & FID \\
      \midrule
      - & FP16 & \textbf{25.48} \\
      \midrule
      Naive & W4A8 & 25.75 \\
      MixDQ & W4A8 & 25.36 \\
      \qsched & W4A8 & \underline{21.41} \\
      \midrule
      Naive & W8A8 & 25.49 \\
      MixDQ & W8A8 & \underline{25.16} \\
      \qsched & W8A8 & 26.34 \\
      \bottomrule
    \end{tabular}
    \caption{Quantized model comparison on SDXL-Turbo under varying bitwidths. FID is computed on the MJHQ dataset.}
    \label{tab:sdxl_turbo}
  \end{subtable}
  \caption{Quantitative evaluation of large-Scale few-step diffusion models with a Stable Diffusion XL backbone.  W4A8 and W8A8 are a $4\times$ and $8\times$ model size reduction in comparison to FP16, yet our method improves over baseline. As FID, FID-SD, and CLIPScore may exhibit reduced reliability at large model scales, we complement these metrics with user preference studies in \cref{fig:pareto}.  }
  \label{tab:sidebyside}
\end{table}

In \cref{tab:sdxl}, we present results on a large-scale 2-step Phased Consistency Model built on the Stable Diffusion XL backbone. \qsched achieves only a 1.2\% FID degradation in the 4W8A setting, demonstrating that quantization-aware preconditioning can maintain high performance even under aggressive compression. In contrast, PTQD fails in this regime, yielding significantly higher FID and visibly poor image quality. We attribute this to its reliance on Gaussian noise assumptions, which break down in few-step diffusion—especially for large models—where each step approximates a segment of the ODE trajectory rather than a Gaussian denoising step. This is expected, given that the true data distribution $p_{data}$ is highly complex and often non-Gaussian.

\paragraph{Results: SDXL-Turbo and FLUX.1[schnell]}
In \cref{tab:sdxl_turbo}, we compare quantization strategies on SDXL-Turbo (4-step inference) using the FID metric on the MJHQ dataset, evaluating two bitwidth settings: W4A8 and W8A8. Under W4A8, \qsched achieves a FID of 21.41, significantly outperforming MixDQ~\cite{zhao2024mixdq} (25.36) and Naive (25.75), demonstrating strong robustness to aggressive quantization. However, at W8A8, \qsched shows a higher FID (26.34) than both MixDQ (25.16) and Naive (25.49), suggesting that its advantages are most pronounced in lower-bit regimes, where other methods degrade more severely.

\begin{wrapfigure}{r}{0.35\textwidth}
  \centering
  \includegraphics[width=0.33\textwidth]{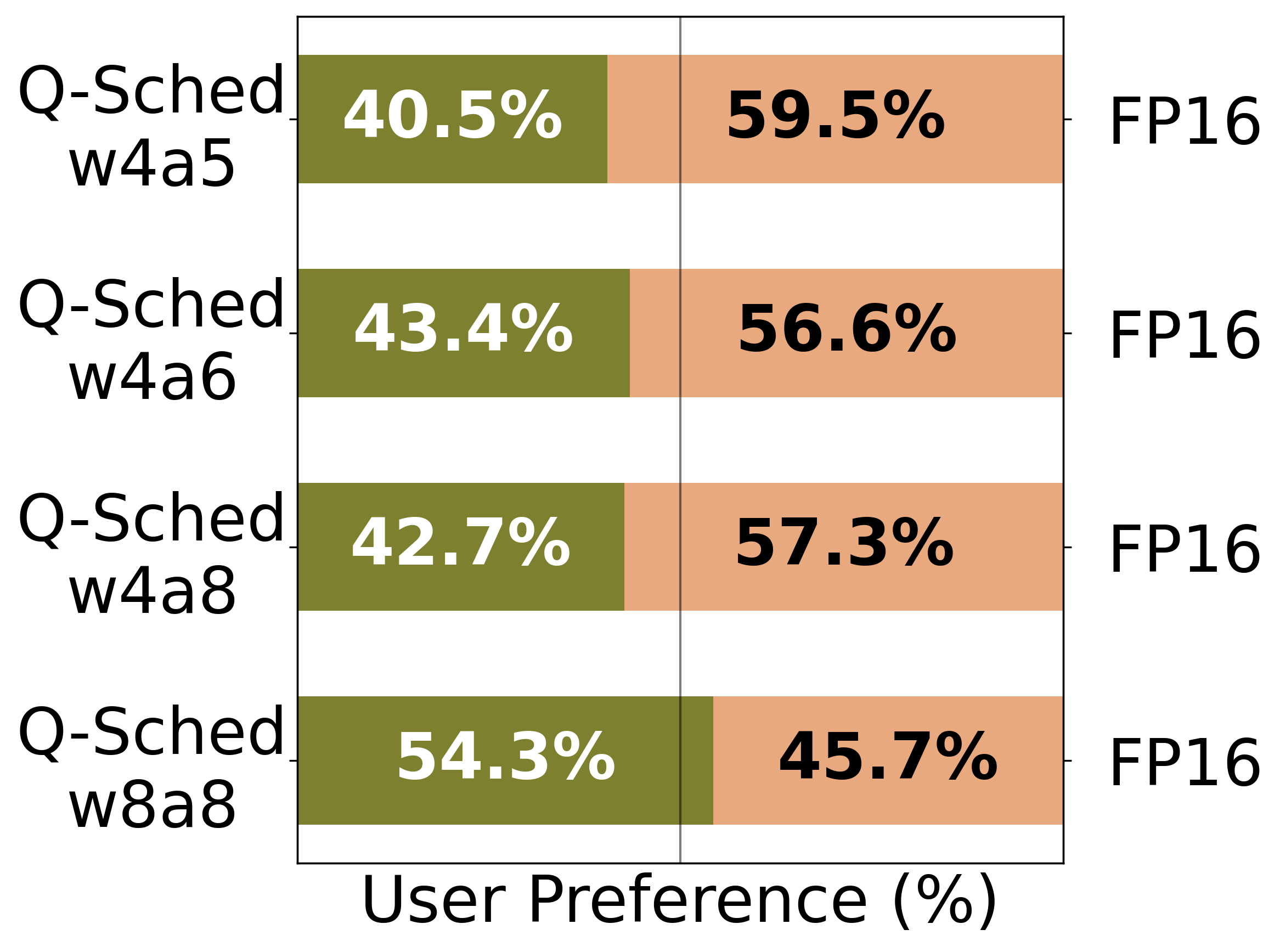}
  \caption{Comparing \qsched across various bit widths.}
  \label{fig:bitwidths}
\end{wrapfigure}

Despite this, a small-scale user study with 52 anonymized participants showed that \qsched was preferred by 56\% of users over MixDQ in the W8A8 setting, indicating that perceptual quality may remain competitive even when FID is less favorable. Additionally, in \cref{fig:pareto}, we present user preference results for \qsched applied to both SDXL-Turbo and FLUX.1 [schnell], showing that it outperforms MixDQ~\cite{zhao2024mixdq} and SVDQuant~\cite{lisvdquant}, respectively, at similar model sizes (see \cref{sec:user_study} for details). We compute an ELO rating, a relative quality ranking inspired by chess scoring, by aggregating all pairwise 1v1 image comparisons across models, where a higher score reflects consistent user preference.

In \cref{fig:bitwidths}, we compare \qsched across bit widths using a user study. W4A4 proved too aggressive, but W4A5 and W4A6 produced images comparable to full precision. 1v1 comparisons with full-precision FLUX.1~\cite{flux1schnell} follow the protocol in \cref{sec:user_study}.

\begin{figure}[htbp]
  \centering

  % --- Figure on the left ---
  \begin{subfigure}[t]{0.58\textwidth}
    \centering
    \includegraphics[width=\linewidth]{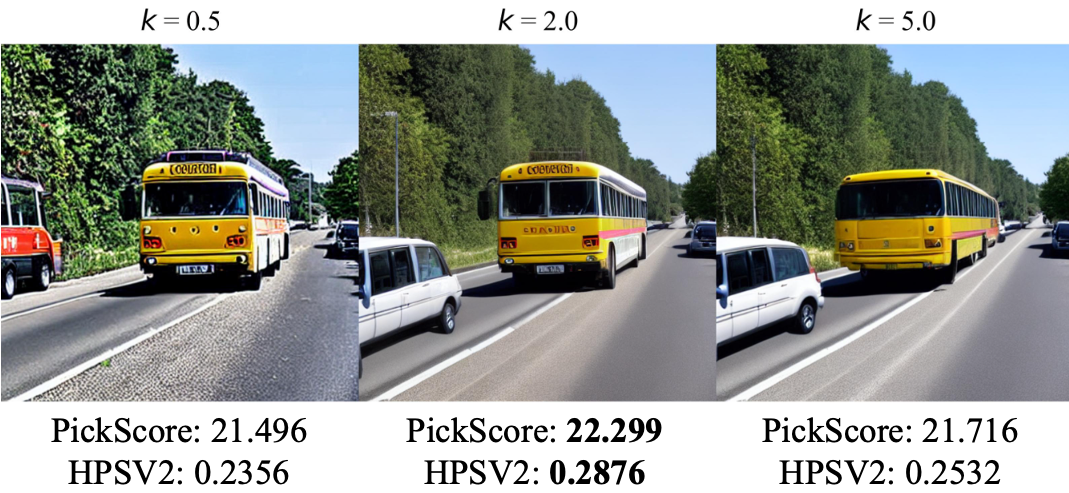}
    \caption{Choice of \textit{k} for the JAQ loss. \textit{k} balances the contribution of $\texttt{TC}(x)$ (CLIPScore) \textit{vs.} $\texttt{IQ}(x)$ (CLIP-IQA-Q). Prompt: "a car and a bus on a french highway".}
    \label{fig:k_choice}
  \end{subfigure}
  \hfill
  % --- Table on the right ---
  \begin{subfigure}[t]{0.39\textwidth}
    \centering
    \small
    \vspace{-11.5em}
    \begin{tabular}{c c c c}
     & $c_\epsilon$ & $c_x$ & ($c_\epsilon$ ,  $c_x$) \\
     \toprule
     PickScore & 21.83 & 22.25 & \textbf{22.30} \\
     HPSV2 & \textbf{0.288} & 0.262 & \textbf{0.288} \\
     JAQ Loss & 3.367 & 3.383 & \textbf{3.392} \\
    \end{tabular}
    \caption{Ablation on choice of pre-conditioning coefficients. We find that optimizing both model and sample coefficients jointly yields optimal image quality. Image quality metrics are averaged over 1024 images generated from the Phased Consistency Models with the SDXL backbone.}
    \label{tab:precond_coeffs}
  \end{subfigure}
  \caption{Ablation studies on various design choices for \qsched.}
  \label{fig:side_by_side}
\end{figure}

\paragraph{Ablation on Pre-Conditioning Coefficients and Loss Function Choice}
We ablate the choice of pre-conditioning coefficients in the Phased Consistency Model by comparing performance when optimizing only the model-side coefficient $c_\epsilon$, the sample-side coefficient $c_x$, or both jointly. As shown in \cref{tab:precond_coeffs}, jointly optimizing both $c_\epsilon$ and $c_x$ consistently yields the best results across all three metrics: PickScore, HPSv2, and JAQ Loss. These findings highlight the importance of treating both denoising and reconstruction terms as tunable components rather than fixing one a priori. All metrics are averaged over 1024 images generated with the SDXL backbone.

\paragraph{How Do We Choose \textit{k} For The JAQ Loss?}

We optimize the \qsched preconditioners using the JAQ loss, which balances image quality and text-image consistency via a tradeoff hyperparameter, $k$. As shown in \cref{fig:k_choice}, small $k$ values can lead to color distortion, while larger values (\textit{e.g.}, $k=5$) cause outputs to drift from the true data distribution. In such cases, the JAQ loss behaves similarly to CLIP-IQA-Q, which lacks sensitivity to concept alignment. We find that a hand-tuned value of $k$ is sufficient for producing a high-quality noise schedule, and the final results are not highly sensitive to its exact choice. Throughout our experiments, we use $k=2$.

For a detailed comparison of loss functions, see \cref{sec:compare_loss}.

\section{Broader Impacts}
Model compression enables wider and more ubiquitous AI usage as we compress large foundation models to run on resource-constrained GPUs. Our work's potential societal consequences are similar to those of prior work as both quantization and few-step diffusions are model compression methods for text-to-image generative models. Generated images have the potential to mislead, mis-represent and cause social harm. We conduct a user preference on a crowd-sourcing platform where generated content is shown to users worldwide and has the potential to cause harm.

\section{Conclusion}
Few-step diffusion models dramatically reduce inference cost by distilling large generative models—such as Stable Diffusion XL—into versions requiring only 2–8 denoising steps, achieving a $5$–$25\times$ speedup. However, these models typically reduce runtime without addressing model size. Our method, \qsched, pushes this efficiency frontier further by introducing quantization into the few-step regime. Through noise-aware preconditioning coefficients, \qsched enables effective quantization with minimal performance loss. We report \textbf{8.0\% and 16.1\% FID improvements} over full-precision baselines for PCMs and LCMs, respectively. A user preference study also shows that \textbf{\qsched outperforms existing quantization methods on FLUX.1[schnell] and SDXL-Turbo in perceived image quality}. These results demonstrate that quantization and few-step distillation are complementary, enabling substantial efficiency gains without compromising generation quality.

\section{Acknowledgments}
This work was supported in part by NSF CCF Grant No. 2107085, iMAGiNE - the Intelligent
Machine Engineering Consortium at UT Austin, and UT Cockrell School of Engineering Doctoral Fellowships. Human Evaluation studies were conducted using Rapidata.

\newpage

\bibliography{references}
\bibliographystyle{plainnat}

\newpage

\appendix

\section{Details on User Preference Assessment}
\label{sec:user_study}

We design our evaluation setup following the user preference study methodology from SDXL-Turbo~\cite{sauer2024adversarial}, with several improvements. For each model pair in this study, we perform 1-vs-1 comparisons based on shared prompts. Human responses, collected via Rapidata~\cite{rapidata2025}, come from evaluators who are presented with two images, each generated by a different model for the same prompt, and are asked: “Which image is of higher quality and more aesthetically pleasing?”

Evaluators are globally sourced and must pass a set of validation questions designed to assess annotation quality. Only those who successfully complete this qualification step are allowed to rate the models.

ELO scores are computed using the same approach as SDXL-Turbo~\cite{sauer2024adversarial}, with K = 32
K=32. We find that this value of K enables more noticeable ranking adjustments, especially when models have similar performance levels.

All models in our study are evaluated using 1,000 prompts sampled from the MJHQ-30k dataset. We release this subset, which we call the \qsched split, to enable consistent benchmarking of future quantization methods. Each prompt is evaluated by four unique annotators. Therefore, each 1-vs-1 comparison results in 4,000 total human annotations.

\section{Compute Resources \& Statistical Significance}
\label{sec:compute}
We conduct all our experiments on a high-end AI server with eight Nvidia A6000s. Each model can be run independently on one A6000 and \qsched takes approximately twenty minutes to run the full grid search.

Our main experiments are averaged over two-three runs but we do not report error bars at this time.

\section{Ablation Studies}

\subsection{Comparing Loss Functions for \qsched}
\label{sec:compare_loss}
\begin{figure}
  \centering
  \begin{subfigure}[t]{0.45\textwidth}
    \centering
    \includegraphics[width=\linewidth]{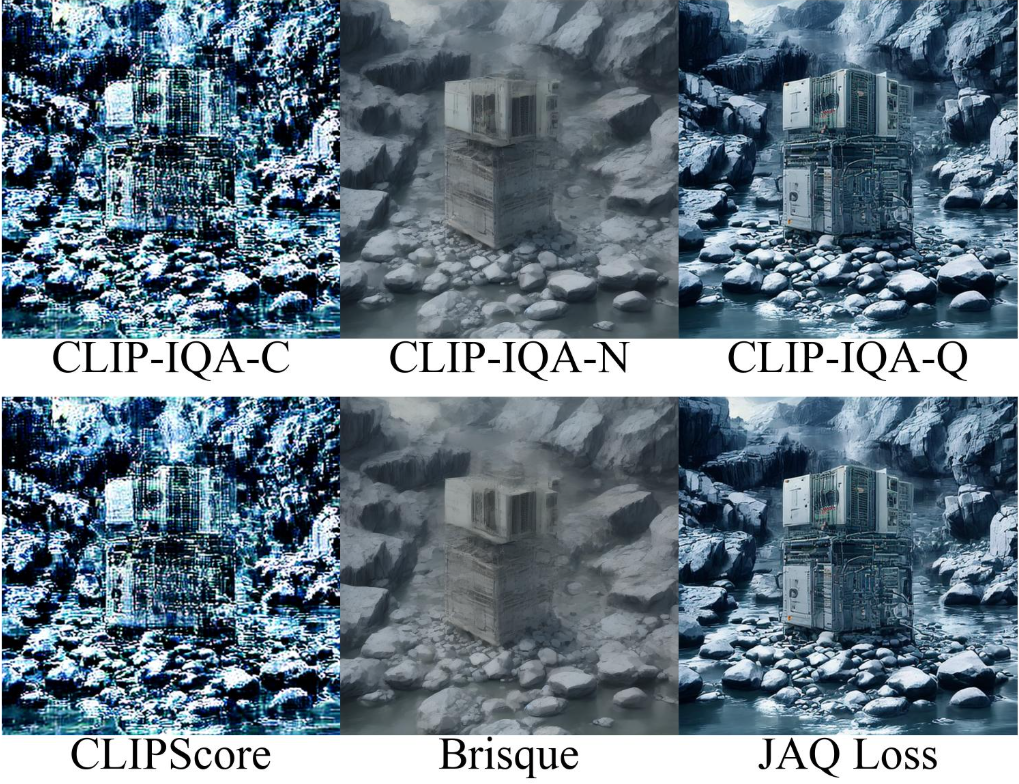}
    \caption{Prompt: "An Eniac computer balanced on top of a stack of rocks over a river"}
    \label{fig:img_quality_2s1cm}
  \end{subfigure}
  \hfill
  \begin{subfigure}[t]{0.45\textwidth}
    \centering
    \includegraphics[width=\linewidth]{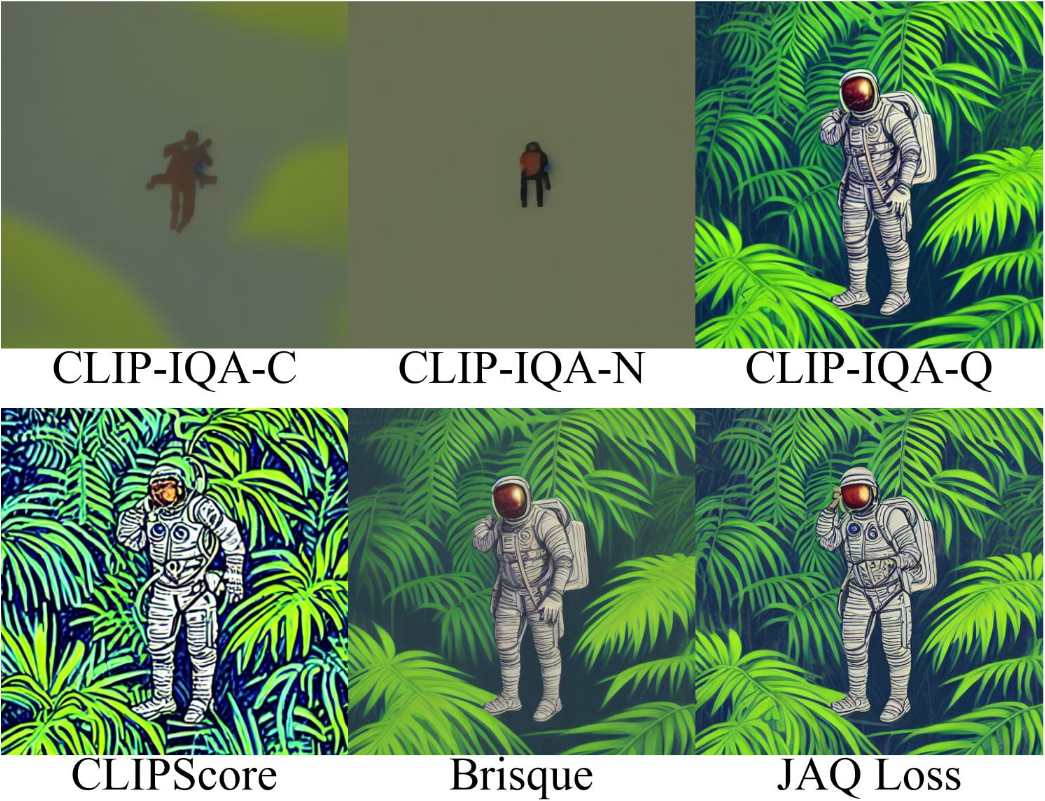}
    \caption{Prompt: "Astronaut in a jungle, cold color palette, muted colors, detailed, 8k"}
    \label{fig:img_quality_eniac}
  \end{subfigure}
  \caption{Optimizing \qsched with various reference-less image quality metrics. Our loss function, JAQ, is a linear combination of CLIPScore and CLIP-IQA-Q. We compare against three CLIP-IQA prompts: Complexity, Noisiness, and Quality denoted as -C, -N, -Q respectively.}
  \label{fig:img_quality}
\end{figure}

To evaluate the overall image quality for text-image generative modeling, CLIPScore~\cite{hessel2021clipscore} is specifically designed to capture text-image compatibility and does not consider overall image quality. In \cref{fig:img_quality}, we illustrate that \qsched optimized with CLIPScore produces an updated noise schedule that is over saturated and lacks image depth. In contrast, Brisque~\cite{brisque} is often used as a standard reference-free image quality metric, but when used in \qsched it creates images with smoother and less detailed features. We consider three variants of CLIP-IQA~\cite{clipiqa} and find that CLIP-IQA using the predefined quality prompt (we denote this version by CLIP-IQA-Q) achieves a noise schedule with high-fidelity images. However, CLIP-IQA-Q has a significant weakness: it cannot properly score images with hallucinations because it does not have an understanding of the underlying image prompt or concept. Therefore, we combine the benefits of CLIPScore and CLIP-IQA-Q into the JAQ loss and find that the resulting schedule fares extremely well with respect to raw image quality as well as to concept adherence.

\subsection{Adding Stochasticity}

Phased Consistency Model's implementation of the original sampler, TCD, is deterministic, meaning that there is no additive noise during sampling. The controllable noise parameter, $\eta$, allows a practitioner to adjust the additive noise during the sampling process and is defined in \cref{eq:sss}. In order to compare PTQD's correction to our method, we ablate across different levels of stochasticity and report performance for six stochasticity levels in \cref{tab:stochasticity}. $\eta = 0$ refers to deterministic sampling and PTQD's uncorrelated noise correction is not used since it adds stochastic noise by construction. Please see the appendix for more details on PTQD's implementation in both deterministic and stochastic sampling regimes.

We find that \qsched outperforms PTQD for all stochasticity regimes on the 2-step phased consistency model. With a simple grid search using our JAQ loss, we can outperform PTQD and the original TCD scheduler in different sampling regimes.

\begin{table}
    \centering
    \caption{Adding Stochasticity and its effect on W4A8 Quantization for PCM using a Stable Diffusion v1-5 backbone. We report FID on COCO-30k. The stochasicity term, $\eta$, controls the amount of added Gaussian noise. $\eta = 0$ is deterministic sampling. }
    \begin{tabular}{c|c c c c c c}
    \toprule
     Method & \multicolumn{6}{c}{$\eta =$} \\
    
         & 0 & 0.1 & 0.3 & 0.5 & 0.7 & 0.9 \\
         \midrule
        TCD & 28.70 & 24.06 & \textbf{23.44} & 22.97 & 26.74 & 22.40 \\
        PTQD & 23.33 & 25.59 & 24.95 & 25.69 & 24.53 & 26.71 \\
        \qsched & \textbf{22.24} & \textbf{19.29} & \textbf{23.44 }& \textbf{19.67} & \textbf{19.46} & \textbf{17.87} \\ 
    \end{tabular}
    \label{tab:stochasticity}
\end{table}

\section{Quantization-Induced Artifacts}
\begin{figure}
    \centering
    \begin{subfigure}[t]{0.32\linewidth}
        \centering
        \caption{Color Distortion}
        \includegraphics[height=2in]{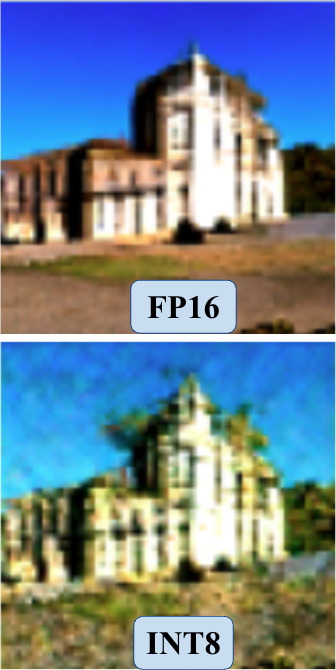}
    \end{subfigure}
    \begin{subfigure}[t]{0.32\linewidth}
        \centering
        \caption{Degradation}
        \includegraphics[height=2in]{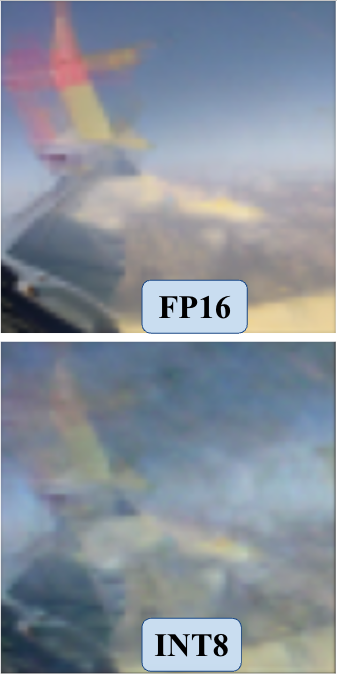}
    \end{subfigure}
    \begin{subfigure}[t]{0.32\linewidth}
        \centering
        \caption{Hallucinations}
        \includegraphics[height=2in]{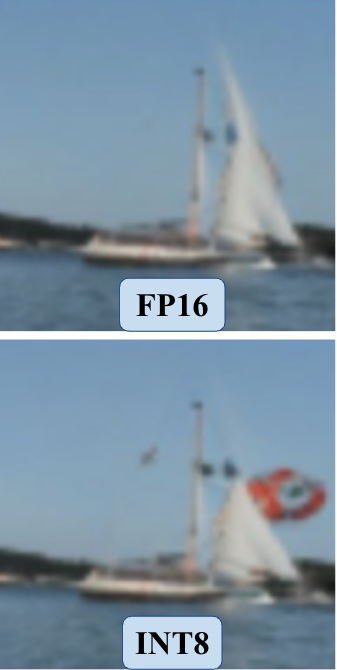}
    \end{subfigure}%
    \caption{Three types of image artifacts that occur when quantizing image generation models. Images are unconditionally generated from a Two-Step Consistency Model~\cite{song2023consistency}.}
    \label{fig:artifacts}
\end{figure}

In our preliminary analysis using a two-step Consistency Model, we observed several characteristic ways in which quantization degrades image quality. As shown in \cref{fig:artifacts}, quantized models tend to exhibit three prominent types of artifacts: color distortion, image degradation, and hallucinated structures. These issues are especially pronounced in low-bit settings and appear consistently across a variety of models and prompts.

\section{Model Size Analysis}

In \cref{tab:model_size}, we show the full model size breakdown for the diffusion model backbone, text encoders, and the VAE decoder. During inference, either one or both text encoders are used, and we do not need the VAE encoder, since this is for training exclusively.

\begin{table}[h]
    \centering
    \caption{FP16 Diffusion Model Size Breakdown (in GB)}
    \begin{tabular}{c|c c c c}
    \toprule
     & LCM & PCM & SDXL-Turbo & FLUX.1[schnell]\\
    \midrule
         UNet/DiT &  1.72 & 4.84 & 1.03 & 4.76 \\
         Text Encoder(s) & 0.25 & 0.29 & 0.33 & 1.95 \\
         VAE Decoder & 0.07 & 0.13 & 0.02 & 0.02 \\
    \midrule
         Total & 2.04 GB & 5.26 GB & 1.37 GB & 6.73 GB\\
    \bottomrule
    \end{tabular}
    \label{tab:model_size}
\end{table}

For our ELO \textit{vs.} Model Size Pareto front in \cref{fig:pareto}, we consider the DiT memory and compute model size by taking the parameter count and multiplying it by the number of bytes required per parameter. For W4A4 + LoRA 64, the setup used for SVDQuant~\cite{lisvdquant}, we compute the number of LoRA parameters using the back-of-the-envelope calculation provided in SVDQuant and add it to this calculation. We provide raw data for clarity in ~\cref{tab:dit_mem}.
\begin{table}[h]
    \centering
    \caption{DiT Memory (in GB) for various bitwidths.}
    \begin{tabular}{ c | c c}
    \toprule
        Precision & SDXL-Turbo & FLUX.1[schnell] \\
        \midrule
         FP16 & 1.03 &  4.76  \\
         W8A8  & 0.51 & 2.38 \\
         W4A4 + LoRA 64 & 0.28 & 1.24 \\
         W4A8 & 0.26 & 1.19 \\
    \bottomrule
    \end{tabular}
    
    \label{tab:dit_mem}
\end{table}

\section{Additional Analysis on COCO-30k}
This result reinforces the core finding of our paper: quantization, when paired with a scheduler designed to account for noise sensitivity (as in \qsched), can be synergistic with few-step diffusion rather than detrimental. Notably, our quantized model achieves a lower FID than the original full-precision model, suggesting that \qsched helps overcome limitations introduced by both step reduction and bit-level compression.

These findings complement the results on SDXL-Turbo and FLUX.1[schnell] discussed in the main paper, and further establish \qsched as a general-purpose solution for high-fidelity, compressed diffusion generation.

\begin{figure}[h]
    \centering
    \includegraphics[height=0.4\linewidth]{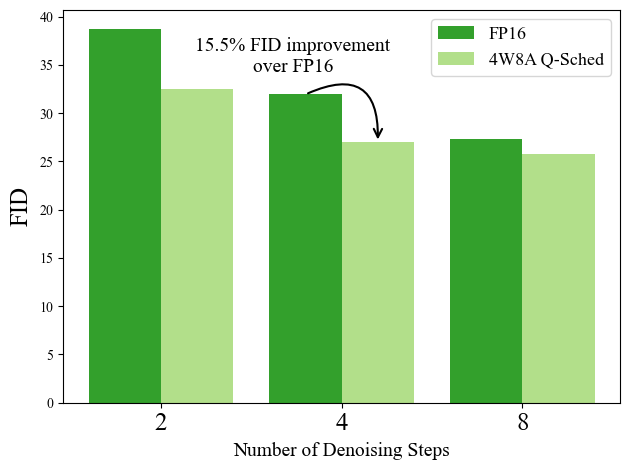}
    \caption{FID on COCO-30k. A 4W8A compressed model with our \qsched scheduler outperforms its FP16 counterpart with a $4\times$ reduction in model size.}
    \label{fig:attn_grabber}
\end{figure}

\section{Applying PTQD to the TCD Scheduler}
\label{sec:ptqd}

Using PTQD's linear parameterization for the quantization error, we substitute $\netq{t} = (1+\gamma) \cdot \eps + \delta $ into \cref{eq:sss}:

\begin{equation}
     \mathbf{x_s} = \frac{\alpha_s}{\alpha_{s'}} \Big(\alpha_{s'}\frac{\mathbf{x_t} - \sigma_t   \net{t}}{\alpha_t} + \sigma_{s'} \net{t}\Big)
     + \frac{\alpha_s}{\alpha_{s'}(1+ \gamma)}(\sigma_{s'} 
 - \frac{\alpha_{s'}\sigma_t}{\alpha_t} ) \delta + \sqrt{1 - \frac{\alpha_s^2}{\alpha_{s'}^2}} \mathbf{z} .
\end{equation}

PTQD assumes the uncorrelated noise is sampled from a normal distribution $\delta \sim N(\mu_\delta, \sigma_\delta)$. This method applies bias correction to handle the mean deviation, $\mu_\delta$, and analytically compute standard deviation, $\sigma_\delta$. We adapt PTQD's approach to the TCD schedule and use the new standard deviation, $\sigma_\delta$ for sampling $\delta$:

\begin{equation}
    \sigma_{\delta}^2 = 1 - \frac{\alpha_s^2}{\alpha_{s'}^2}(1 -  (\frac{\delta (\sigma_{s'} 
 - \frac{\alpha_{s'}\sigma_t}{\alpha_t})}{(1+ \gamma)})^2) .
\end{equation}

For the edge case, where $\sigma_\delta < 0$, the deviation is set to zero ($\sigma_\delta = 0$). The proof for extending PTQD to the TCD scheduler is in the appendix. 

PTQD attempts to model the distribution shift from a full precision to quantized model using two assumptions:

\begin{enumerate}
    \item The quantized model's distribution shift can be modeled through a linear correction term.
    \item The uncorrelated quantization noise is normally distributed.
\end{enumerate}

While these assumptions are similar to prior work on diffusion models, they are likely to break down on the few-step diffusions where the denoising process is distilled from many steps and is not expected to be linear nor follow a Gaussian distribution. 

\paragraph{Quantization Noise Correction using PTQD}

Based on the PTQD quantization noise assumption, the quantization error is linearly parametrized as
$\Delta \eps = \gamma \cdot \eps + \delta$ where $\gamma,\delta$ are learnable parameters corresponding to the correlated noise w.r.t. full precision and the uncorrelated noise respectively. PTQD models the uncorrelated noise as Gaussian (\textit{i.e.}, $\delta \sim \mathcal{N}(\mu_q, \sigma_q)$).

\paragraph{Variance Schedule Calibration for Trajectory Consistency Distillation (TCD)}

TCD's Strategic Stochastic Sampling (SSS) using a quantized network $\netq{t}$ is given by:

\begin{equation}
    \mathbf{x_s} = \frac{\alpha_s}{\alpha_{s'}} \Big(\alpha_{s'}\frac{\mathbf{x_t} - \sigma_t \netq{t}}{\alpha_t} + \sigma_{s'}\netq{t} \Big) + \sqrt{1 - \frac{\alpha_s^2}{\alpha_{s'}^2}} \mathbf{z}
\end{equation}

Using PTQD's linear parametrization for the quantization error, we substitute $\netq{t} = (1+ \gamma) \cdot \eps + \delta $:

\begin{align}
    \mathbf{x_s} &= \frac{\alpha_s}{\alpha_{s'}} \Big(\alpha_{s'}\frac{\mathbf{x_t} - \sigma_t ( (1+ \gamma) \cdot \net{t} + \delta)}{\alpha_t} + \sigma_{s'}((1+ \gamma) \cdot \net{t} + \delta )\Big) + \sqrt{1 - \frac{\alpha_s^2}{\alpha_{s'}^2}} \mathbf{z} \\
    &= \frac{\alpha_s}{\alpha_{s'}} \Big(\alpha_{s'}\frac{\mathbf{x_t} - \sigma_t (1+ \gamma)  \net{t}}{\alpha_t} + \sigma_{s'}(1+ \gamma) \net{t} - \frac{\alpha_{s'} \sigma_t \delta}{\alpha_t} + \sigma_{s'} \delta\Big) + \sqrt{1 - \frac{\alpha_s^2}{\alpha_{s'}^2}} \mathbf{z} \\
    &= \frac{\alpha_s}{\alpha_{s'}} \Big(\alpha_{s'}\frac{\mathbf{x_t} - \sigma_t (1+ \gamma)  \net{t}}{\alpha_t} + \sigma_{s'}(1+ \gamma) \net{t}\Big) + \frac{\alpha_s}{\alpha_{s'}}(\sigma_{s'} 
 - \frac{\alpha_{s'}\sigma_t}{\alpha_t} )\delta + \sqrt{1 - \frac{\alpha_s^2}{\alpha_{s'}^2}} \mathbf{z} \\
\end{align}

The correlated noise can be corrected by applying:
\begin{align}
    \frac{\netq{t}}{1+ \gamma} &= \frac{(1+ \gamma) \net{t} + \delta}{1+ \gamma} \\
    &= \net{t} + \frac{\delta}{1+ \gamma}
\end{align}

The resultant SSS sampling step becomes:
\begin{align}
    \mathbf{x_s} &= \frac{\alpha_s}{\alpha_{s'}} \Big(\alpha_{s'}\frac{\mathbf{x_t} - \sigma_t   \net{t}}{\alpha_t} + \sigma_{s'} \net{t}\Big) + \frac{\alpha_s}{\alpha_{s'}(1+ \gamma)}(\sigma_{s'} 
 - \frac{\alpha_{s'}\sigma_t}{\alpha_t} )\delta + \sqrt{1 - \frac{\alpha_s^2}{\alpha_{s'}^2}} \mathbf{z} \\
\end{align}

The variance schedule becomes:

\begin{align}
 %    (\frac{\alpha_s}{\alpha_{s'}(1+ \gamma)}(\sigma_{s'} 
 % - \frac{\alpha_{s'}\sigma_t}{\alpha_t} ))^2 \delta^2 + \sigma_{*}^2 &= 1 - \frac{\alpha_s^2}{\alpha_{s'}^2} \\
 \sigma_{\delta}^2 &=  1 - \frac{\alpha_s^2}{\alpha_{s'}^2} -  (\frac{\alpha_s}{\alpha_{s'}(1+ \gamma)}(\sigma_{s'} 
 - \frac{\alpha_{s'}\sigma_t}{\alpha_t} ))^2 \delta^2 \\
 &= 1 - \frac{\alpha_s^2}{\alpha_{s'}^2}(1 -  \frac{(\sigma_{s'} 
 - \frac{\alpha_{s'}\sigma_t}{\alpha_t} )^2}{(1+ \gamma)^2} \delta^2)  \\
  &= 1 - \frac{\alpha_s^2}{\alpha_{s'}^2}(1 -  (\frac{\delta (\sigma_{s'} 
 - \frac{\alpha_{s'}\sigma_t}{\alpha_t})}{(1+ \gamma)})^2) 
\end{align}

Since $\mathbf{z}\sim N(\mu_\delta, \sigma_\delta)$, we must handle the edge case when $\sigma_\delta < 0$. If the variance is negative, we simply set $\sigma_\delta = 0$.

Upon comparing \qsched to PTQD you may ask "\textit{Why is \qsched able to learn a better noise schedule when it is also a linear correction?}" \qsched learns scalar coefficients on $x_t$ and $\eps$ that are optimized with respect to the reference-free JAQ loss. This allows us to learn a new schedule with linear corrections to improve our overall noise schedule, rather than matching the existing full precision schedule. This is an important distinction from PTQD, which tries to learn a linear correction with respect to full precision, which may not be possible since quantization produces a nonlinear distortion on the diffusion model. In short, PTQD attempts to match the full precision sampling trajectory, whereas \qsched aims to learn a new sampling trajectory given a compressed $\eps$.

\section{Strict Guarantees for Quantization-Aware Scheduling}
Let us consider the few-step sampling trajectories for the pre-trained and quantized models, parametrized by $\mathcal{E}_\theta(t)$ and $\mathcal{E}^Q_\theta(t)$ respectively. These two few-step diffusion models sample at the same time-steps,  $0 = t_0 < t_1, t_2 \cdots t_N = T$,  where $N$ represents the number of steps in the few-step model. For ease of notation, we will use the time-step 0 to refer to $t_0$ and 1 to refer to $t_1$, etc.  A denoising step going from time $t+1 \rightarrow t$, produces a partially denoised image, $x_{t}$, and its quantized counterpart, $x^Q_{t}$. Following directly from Equation 9, the denoising error, $\Delta x_t = x_t - x^Q_t$, can be explicitly computed as:

\begin{align}
    \Delta x_t  &= \frac{\alpha_t}{\alpha_{t'}} \Big(\alpha_{t'}\frac{ \Delta x_{t+1} - \sigma_{t+1} (\mathcal{E}_\theta(t+1) - \mathcal{E}^Q_\theta(t))}{\alpha_{t+1}} + \sigma_{t'}(\mathcal{E}_\theta(t+1) - \mathcal{E}^Q_\theta(t+1) ) \Big)  \\
    &=  \frac{\alpha_t}{\alpha_{t+1}} \Delta_{t+1} + \frac{\alpha_t}{\alpha_{t'}}(\sigma_{t'} - \frac{\sigma_{t+1}}{\alpha_{t+1}})(\mathcal{E}_\theta(t+1) - \mathcal{E}^Q_\theta(t+1))  \\
    &= k_{t} \Delta x_{t+1} + m_{t} \Delta \mathcal{E}_\theta(t+1))
\end{align}

where we define the sampler coefficients as $k_t = \frac{\alpha_t}{\alpha_{t+1}}$, $m_t = \frac{\alpha_t}{\alpha_{t'}}(\sigma_{t'} - \frac{\sigma_{t+1}}{\alpha_{t+1}})$ and denote the change in the network as $\Delta \mathcal{E}_\theta(t) = \mathcal{E}_\theta(t) - \mathcal{E}^Q_\theta(t)$. Assuming the initial denoised image is the same ($x_N = x^Q_N$), the error in the final denoised image, $\Delta x_0$, is given by:

\begin{align}
    \Delta x_0 &= k_0\Delta x_1 + m_0 \Delta \mathcal{E}_\theta(1) \\
    % &= k_0k_1k_2(k_3 \Delta x_4 + m_3 \Delta \mathcal{E}_\theta(4}) + k_0k_1m_2 \Delta \mathcal{E}_\theta(3) +  k_0m_1 \Delta \mathcal{E}_\theta(2) + m_0 \Delta \mathcal{E}_\theta(1}) \\
    &= k_0k_1k_2...(k_N \Delta x_N + m_{N-1} \Delta \mathcal{E}_\theta(N)) +\dots + k_0k_1m_2 \Delta \mathcal{E}_\theta(3) +  k_0m_1 \Delta \mathcal{E}_\theta(2) + m_0 \Delta \mathcal{E}_\theta(1) \\
    &= \sum_{s=1}^S \Big( \Pi_{v=0}^{s-2} k_v \Big) m_{s-1} \Delta \mathcal{E}_\theta(s) 
\end{align}

\subsection{Expected Quantization Error}
The average expected error over all images in a given dataset, $x_0 \in \mathcal{D}$, is given by:

\begin{align}
    E[||\Delta x_0||]  &= \sum_{s=1}^S \Big( \Pi_{v=0}^{S-2} k_v \Big) m_{s-1} E[||\Delta \mathcal{E}_\theta(s)||] 
    \label{eq:expectation}
\end{align}

since $E[||\Delta x_0||]$ is a homogeneous function.

In Q-Sched, we apply our quantization-aware pre-conditioning  on every noise coefficient: $\Tilde{m}_t = c^\epsilon_t \cdot m_t$ and $\Tilde{k}_t = c^x_t \cdot m_t$ . Let us denote the expected error induced by Q-Sched with respect to the pre-trained model's $x_0$ as $E[||\Delta \Tilde{x_0}||]$. 

We empirically show in Tables 1 and 2 that $E[||\Delta x_0 ||] \neq 0$ since the images produced by the naive quantization method produce a different FID from the original pre-trained model's image distribution.  Since Equation \ref{eq:expectation} is a linear function of  $k_t, m_t, \forall  t \in 1 \dots N $, and there is a global minimum at $E[||x_0 - x_0||] = 0$, it must be that $\exists \Tilde{m}^*_t, \Tilde{k}^*_t \forall t $ such that $ E[||\Delta \Tilde{x_0}||] < E[||\Delta x_0||] $. In short, we guarantee that there exists quantization-aware coefficients that  strictly improve our expected quantization error over naive quantization.

\subsection{Aside: Positive Sampler Coefficients}
The TCD Scheduler has $\beta_0 = 0.0085, \beta_N = 0.012$, $\alpha_t = 1 - \beta_t, \sigma_t = \Pi_{i = 0}^{t} \alpha_i$ with a scaled linear schedule:

\begin{equation}
    \beta_t = \Big(\sqrt{\beta_0} + t \cdot (\sqrt{\beta_N} - \sqrt{\beta_0})\Big)^2
\end{equation}

Therefore:
$1 > \alpha_0 > \alpha_1 > \dots > \alpha_N > 0$ and $1 > \sigma_0 > \sigma_1 > \dots \sigma_N > 0$. We note the $t' = (1-\gamma) t$ where $\gamma \in [0,1]$, so $t' \leq t$. This implies that $\sigma_{t'} > \sigma_{t+1}$ so:

\begin{equation}
    k_t > 0 \hspace{10mm},\hspace{10mm} m_t = \frac{\alpha_t}{\alpha_{t'}} (\sigma_{t'} - \frac{\sigma_{t+1}}{\alpha_{t+1}}) > 0 
\end{equation}

This illustrates that $k_t, m_t \in \mathbb{R}^+$.

\section{Qualitative Analysis}
\label{sec:good_v_bad_prompts}
\begin{figure}
  \centering
  \begin{subfigure}[t]{0.45\textwidth}
    \centering
    \includegraphics[height=0.75\textheight]{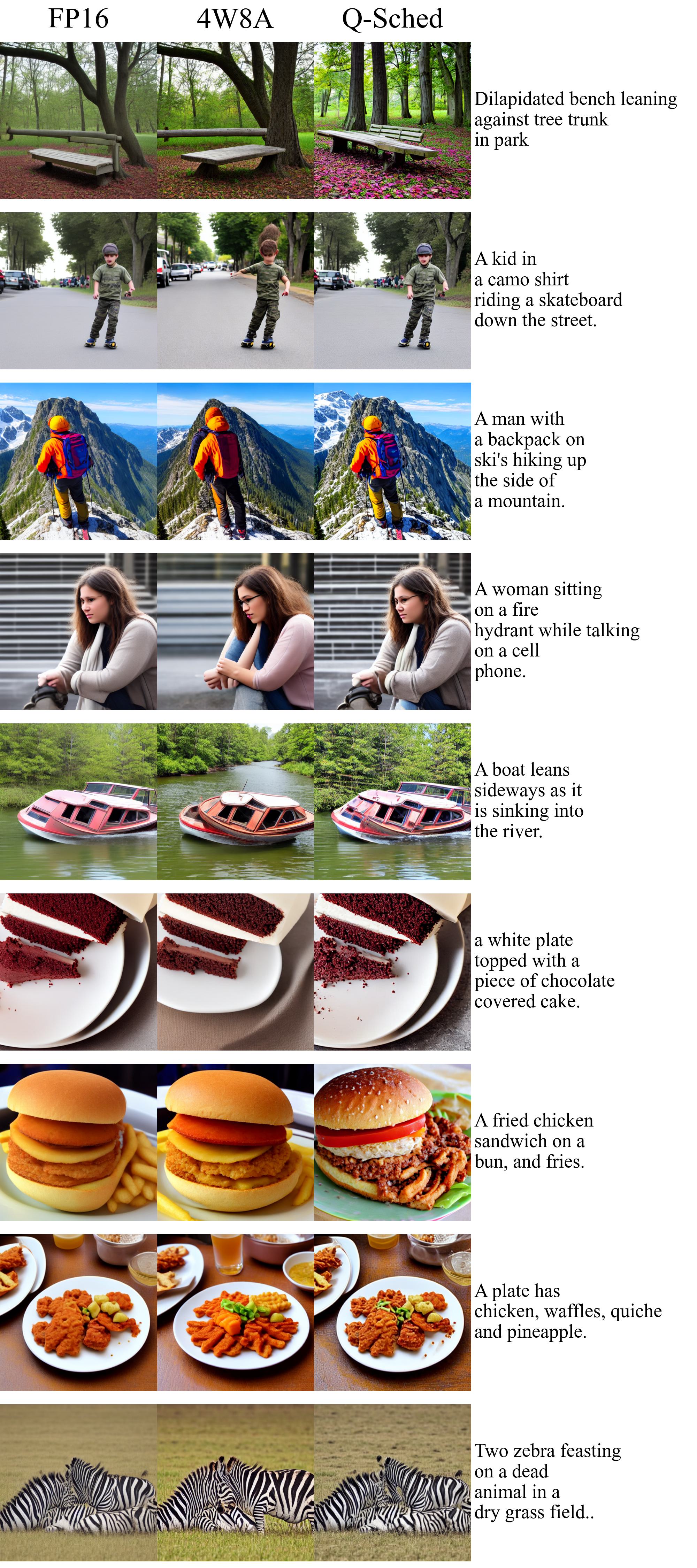}
  \end{subfigure}
  \hfill
  \begin{subfigure}[t]{0.45\textwidth}
    \centering
    \includegraphics[height=0.6\textheight]{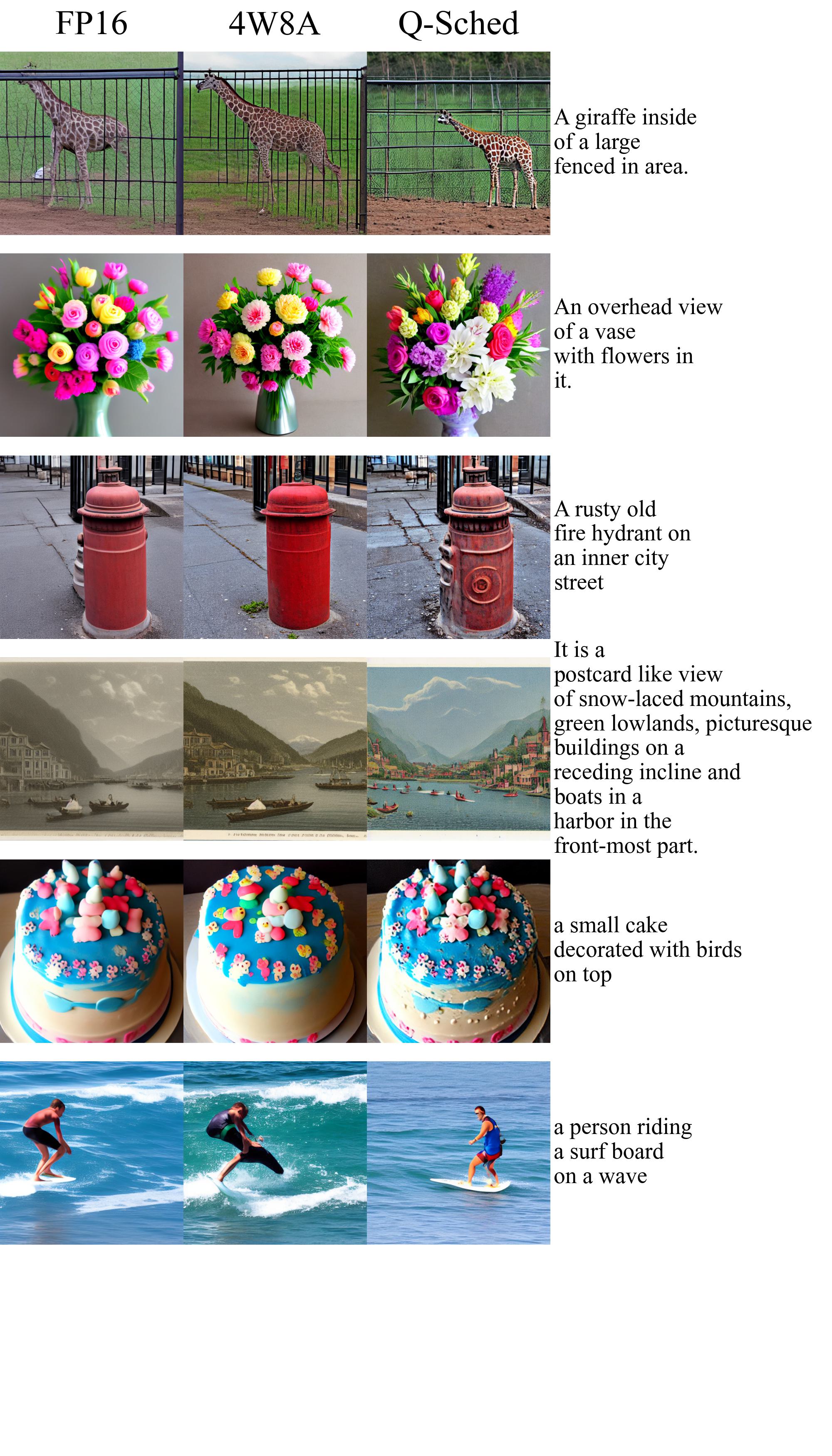}
  \end{subfigure}
  \caption{Selected prompts where \qsched outperforms full precision.}
  \label{fig:good_captions}
\end{figure}

\begin{figure}
    \centering
    \includegraphics[width=0.65\linewidth]{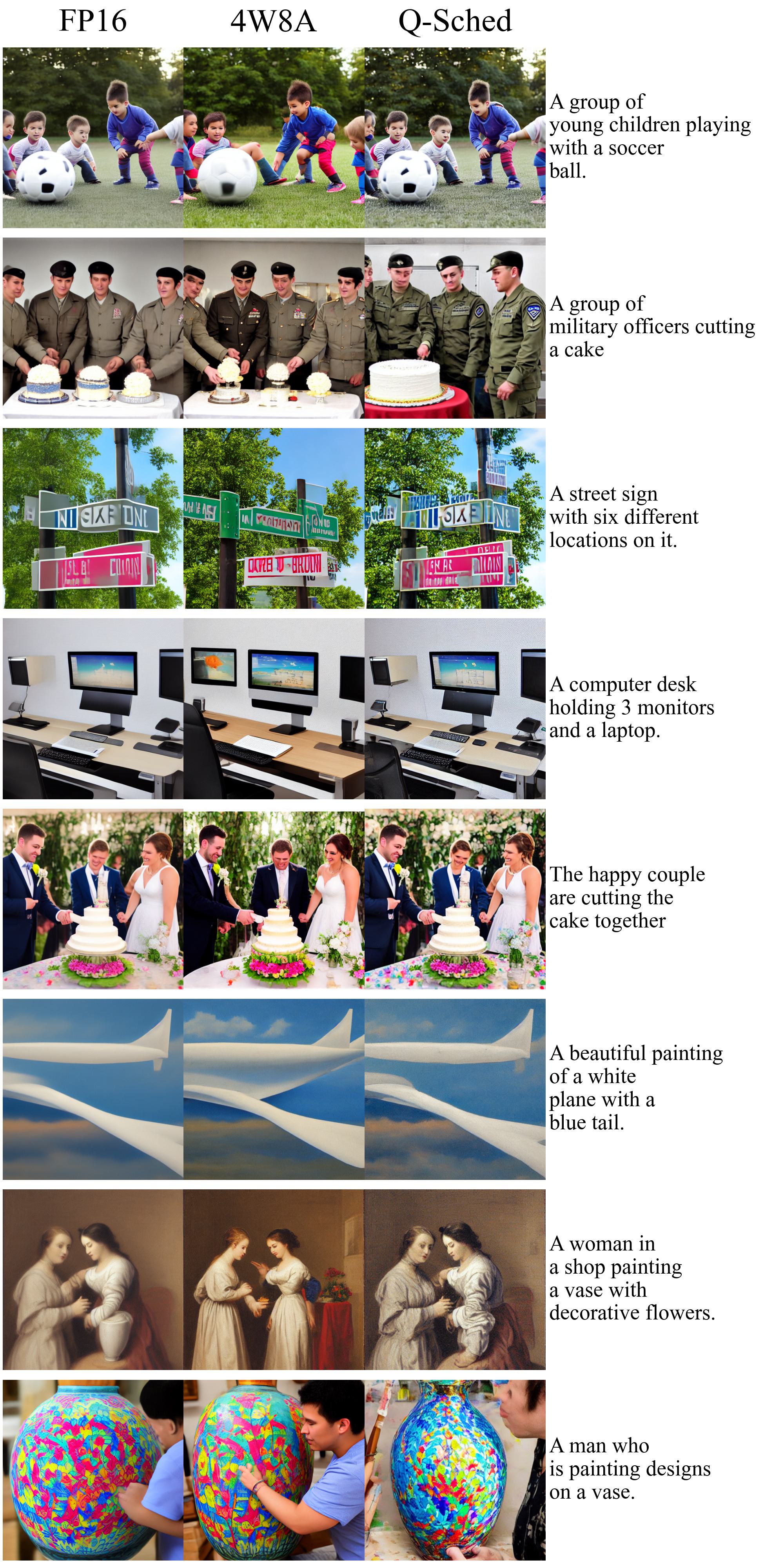}
    \caption{Selected prompts where \qsched fails to improve over 4W8A.}
    \label{fig:bad_captions}
\end{figure}

We present qualitative results from a few hand-picked examples of good and bad results for our method in \cref{fig:good_captions} and \cref{fig:bad_captions} respectively. Good examples were easier to find than bad examples, as our optimized \qsched scheduler tends to provide more detail and an enhanced version of the 4W8A model. In particular, we see additional generative details in the first and third prompts in \cref{fig:good_captions}, where our method generates fine-grained details that are not present in either full precision or 4W8A. We also note that our method provides additional texture such as on the fried chicken sandwich on seventh prompt and more realistic hair textures in the fourth prompt.

Our method still struggles with multi-person images such as the first and second prompts in \cref{fig:bad_captions}. We note that plain text is not improved with our method, as shown in the third prompt. We also find that some artifacts are not rectified with our model, such as in the sixth prompt where the plane silhouette is not properly generated. We also notice a slight divergence in some images from the original prompt such as in the seventh prompt, where a vase is not present in our generated image. Overall, these bad examples are fewer than the good examples and our significant FID improvement shows the \qsched is capable of providing very solid results.

\end{document}